\documentclass[runningheads]{llncs}

\usepackage{eccv}

\usepackage{eccvabbrv}

\usepackage{graphicx}
\usepackage{multirow}
\usepackage{booktabs}
\RequirePackage{xspace}
\RequirePackage{graphicx}
\RequirePackage{amsmath}
\RequirePackage{amssymb}
\RequirePackage[numbers,sort&compress]{natbib}
\RequirePackage{bm}

\usepackage[accsupp]{axessibility}  

\usepackage[pagebackref,breaklinks,colorlinks,citecolor=eccvblue]{hyperref}

\usepackage{orcidlink}

\begin{document}

\title{Baking Relightable NeRF for Real-time Direct/Indirect Illumination Rendering} 

\titlerunning{Abbreviated paper title}

\author{Euntae Choi\inst{1} \and
Vincent Carpentier \inst{1} \and
Seunghun Shin \inst{1} \and
Sungjoo Yoo \inst{1}}

\authorrunning{F.~Author et al.}

\institute{Department of Computer Science and Engineering, Seoul National University, South Korea
}

\maketitle

\begin{abstract}
  Relighting, which synthesizes a novel view under a given lighting condition (unseen in training time), is a must feature for immersive photo-realistic experience.
However, real-time relighting is challenging due to high computation cost of the rendering equation which requires shape and material decomposition and visibility test to model shadow. Additionally, for indirect illumination, additional computation of rendering equation on each secondary surface point (where reflection occurs) is required rendering real-time relighting challenging. 
We propose a novel method that executes a CNN renderer to compute primary surface points and rendering parameters, required for direct illumination. We also present a lightweight hash grid-based renderer, for indirect illumination, which is recursively executed to perform the secondary ray tracing process. Both renderers are trained in a distillation from a pre-trained teacher model and provide real-time physically-based rendering under unseen lighting condition at a negligible loss of rendering quality.
\end{abstract}

\section{Introduction}
\label{sec:intro}
Baking is one of the most effective methods to boost the speed of neural rendering and there have been active studies~\cite{plenoctrees,snerg,r2l,mr2l,Wan_2023_CVPR,bakedsdf,rojas2023rerend}. The basic philosophy of baking is to utilize distillation from pre-trained neural rendering models~\cite{nerf,tensorf} to a baked model that usually comes with lightweight architecture and efficient rendering pipeline. As result, the number of per-ray sampled points~\cite{r2l,mr2l,Wan_2023_CVPR,bakedsdf} and the cost of per-sample point computation~\cite{plenoctrees,snerg,rojas2023rerend} are drastically reduced.

Neural relighting~\cite{nerv,nerd,neuralpil,physg,nerfactor,invrender,tensoir,nvdiffrecmc} is a critical problem that seeks to exploit inverse rendering using deep learning to realize more realistic rendering and wide applicability. Recent solutions introduce an encoder-decoder structure that takes modeled primary surface points from an SDF model-based backbone as input for BRDF estimation~\cite{invrender}
or utilize voxel grids to implicitly predict in the feature space~\cite{tensoir}. 
However, as experiments have shown, optimized codes still take more than 1 second rendering a single frame, making real-time applications infeasible. 
It is because neural relighting requires more complex and computationally heavy models since it needs to perform material and shape decomposition, e.g., BRDF and secondary ray casting (for shadow and indirect illumination modeling) on top of the vanilla neural rendering. 

Especially, for more realistic relighting, indirect illumination must also be rendered. This requires modeling the coordinates of the secondary bounce point, necessitating not just a simple hit test but also the calculation of the expected depth for the shadow ray, and additionally performing recursive BRDF estimation and hit test at secondary surface points. Thus, indirect illumination requires much larger computation cost.
Hence, existing models improve material and shape decomposition by modeling indirect illumination at train time but tend to perform only direct illumination in test time evaluation.

To the best of our knowledge, there has been no baking method proposed for relightable neural rendering models, especially one that handles indirect illumination at test time. This paper proposes a two-fold solution to address these challenges. Our core idea combines a CNN-based renderer inspired by the neural light field (NeLF)~\cite{r2l} and GAN~\cite{stylegan2} to efficiently compute all quantities required for direct illumination, followed by a multi-resolution hash grid-based encoder-decoder model for hit testing and indirect illumination rendering.

Given a camera pose, our CNN-based direct illumination renderer predicts BRDF parameters and surface coordinates. This approach integrates the efficient residual MLP network from MobileR2L~\cite{mr2l} with a super-resolution stage and employs a bilinear upsampling layer for smooth output in conjunction with progressive growing and output skip. As a result, all computations necessary for the primary ray are completed in a single forward pass of the neural network.

Additionally, we propose a hash grid-based renderer that handles the visibility and second recursion calculation of the rendering equation, advancing the quality of direct illumination (by modeling shadow) and enabling accelerated indirect illumination rendering (for reflection modeling). This model encompasses a hash grid encoder that outputs features at each surface point, a BRDF decoder that predicts material and surface normal from these features, and an implicit ray tracer (IRT) that determines visibility in the direction of secondary rays and the expected depth (in the case of shadow rays). The secondary ray tracing uses the surface coordinates from our proposed direct illumination renderer and depth output from IRT to obtain secondary surface coordinates. By recursively executing the hash renderer at these points to evaluate the rendering equation, indirect illumination is realized. Like the direct illumination case, since the entire process consists of only neural network forward passes, rapid relighting becomes feasible.

In summary, we propose a model that uses the CNN renderer to compute primary surface points and rendering parameters, followed by a recursive operation of the hash renderer to perform the secondary ray tracing process, effectively baking the entire relighting workflow. Experimental results on a total of 6 synthetic scenes under unseen light condition have shown that compared to our selected teacher model, our method achieves up to \textbf{84.62} times faster direct illumination rendering with an average PSNR loss of 0.62 for TensoIR-synthetic and a gain of 0.37 for InvRender-synthetic, and for combined direct/indirect illumination, up to \textbf{85.4} times faster rendering speed with an average PSNR loss of 0.39 and a gain of 0.27, respectively. But in terms of SSIM and LPIPS, ours can closely match or outperform the teacher model overall.

\section{Preliminaries}
\label{sec:prelim}

\subsection{Baking NeRF}

Baking has been actively studied since the introduction of the original NeRF method~\cite{nerf}. 
In \cite{plenoctrees,snerg}, the authors propose distilling a pre-trained NeRF model into a voxel grid which contains pre-computed results at each voxel thereby reducing computation cost in rendering time. They also reduce per-ray computation cost by exploiting sparsity (by skipping computation in empty space) and visibility (avoiding computation at unseen sampling points).
In \cite{Wan_2023_CVPR,bakedsdf,rojas2023rerend}, the authors propose building meshes from the pre-trained NeRF models. Specifically, in \cite{Wan_2023_CVPR}, dual meshes are proposed to cover the surface of target object and, at a ray-triangle intersection, a small MLP is executed to produce the ray color from the features of the dual mesh. In \cite{rojas2023rerend}, a single mesh is utilized and, at the ray-triangle intersection, the pixel color is computed with a very small computation cost of three vector dot products thereby offering a rendering speed of up to 1000 frames per second.

Light field network based baking methods are the most related to our method. 
In \cite{r2l}, a neural light field (NeLF) model is proposed to minimize the frequency of executing the MLP model. For each ray, the model is executed only once with all the sample point coordinates as the input. In \cite{mr2l}, the authors further improve the NeLF model by judiciously exploiting super-resolution thereby reducing the number of input pixels which leads to faster rendering.
To the best of our knowledge, there is no previous work on baking for relighting.
Our work contributes to applying the baking approach to relighting.

\begin{figure*}[t] 
\centering
\includegraphics[width=0.8\textwidth]{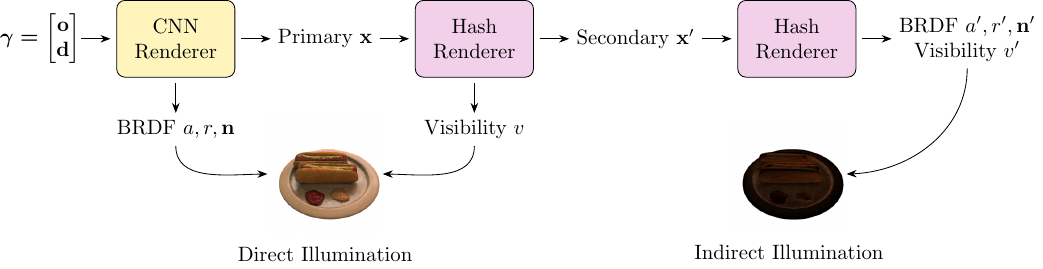}
\caption{Overview of the rendering process of our proposed baked model.}\label{fig:overview}
\end{figure*}

\subsection{Monte Carlo Lighting with Indirect Illumination}

\begin{equation} \label{eqn:renderingeqn}
L(\bm{x}, \bm{\omega}_o) \approx \frac{1}{N} \sum_{i=1}^{N} \frac{L_i(\bm{x}, \bm{\omega}_i)V(\bm{x}, \bm{\omega}_i)f_r(\bm{x}, \bm{\omega}_i, \bm{\omega}_o)(\bm{\omega}_i\cdot\textbf{n})}{p(\bm{\omega}_i)}
\end{equation}

Recent relightable NeRF models are adopting Monte Carlo (MC) estimation-based rendering pipeline \cite{tensoir,nvdiffrecmc}. While the computational load may be heavier and the results are noisier compared to approximated techniques such as Spherical Gaussian (SG)~\cite{invrender,physg} or pre-integration methods~\cite{neuralpil}, these models allow for adjustable sample-per-pixel (spp) values, enabling a balance between quality and speed. Moreover, they can model visibility with greater precision, generally resulting in better rendering quality.

Since a lot of previous works adopt image-based lighting with equirectangular environment maps where each pixel is considered as a point light source, this paper also adopts this setting. The 1-bounce rendering (direct illumination) process is as follows: we first obtain the material parameters and surface normal for the current pixel and then compute the BRDF and cosine factor. After sampling the incoming light direction $\bm{\omega}_i$, the pixels of the environment map are interpolated to calculate the incoming radiance $L_i$. We determine the probability distribution function (pdf) for each direction from the sampling distribution of choice and compute the rendering equation as given in Eqn. \ref{eqn:renderingeqn}.


To account for shadows, we model the visibility of incoming light based on the scene's geometry, excluding the light corresponding to shadow rays from computation. 
In our proposed baking method, we support teacher models based on mesh, SDF and voxel grid.
For mesh-based scenes, traditional ray tracing engines such as OptiX are used to perform hit tests. In the case of SDF-based models~\cite{invrender}, in-house sphere tracing is implemented to produce ground truth visibility and train an MLP to match it. With voxel grid-based models~\cite{tensoir}, volume rendering is executed from surface points in the direction of incoming light to calculate transmittance values, which are then used as visibility.

For more realistic results, the effects of multiple-bounce rays must be computed. This requires recursively calculating the rendering equation, and at each bounce point beyond the second, BRDF computations and light direction sampling must be performed. For simplicity, this explanation is limited to 2-bounce rendering. We calculate the coordinates of the 2nd bounce point for the shadow ray determined during direct illumination using ray tracing. We obtain the BRDF parameters for that point, sample the incoming light direction again, compute the visibility to eliminate shadow rays for each direction, and then compute the rendering equation.

\begin{figure*}[ht]
     \centering
     \begin{subfigure}[b]{0.55\textwidth}
         \centering
         \includegraphics[width=\textwidth]{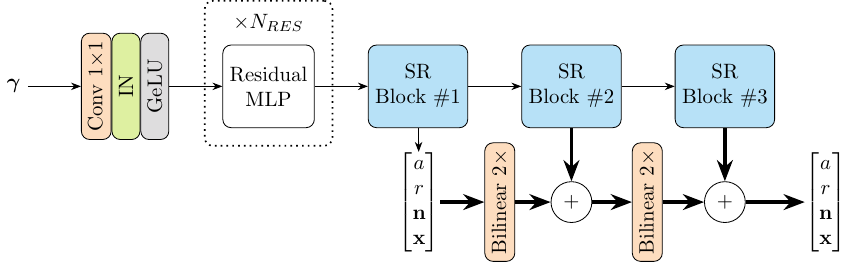}
         \caption{}
         \label{fig:cnn_a}
     \end{subfigure}
     \hfill
     \begin{subfigure}[b]{0.35\textwidth}
         \centering
         \includegraphics[width=\textwidth]{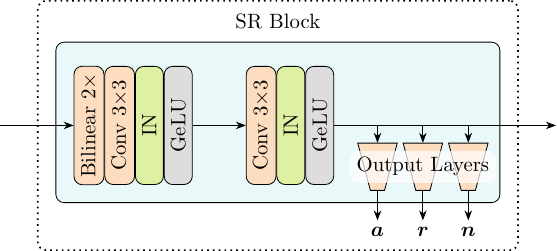}
         \caption{}
         \label{fig:cnn_b}
     \end{subfigure}
   \caption{Architecture of our proposed CNN-based renderer. Instance normalization \cite{in} with affine parameters and GELU \cite{gelu} are used for all normalization and activation layers, respectively. \textbf{(a)}: Our direct illumination renderer consists of an initial 1x1 conv-norm-act layer, a deep residual MLP, and three super-resolution blocks. Each output from an SR block is bilinearly upsampled and added to the output from the next SR block. \textbf{(b)}: Inside an SR block, an input feature map is upsampled bilinearly and processed with two consecutive 3x3 conv-norm-act layers. Then, the output feature map is fed to the output mapping layers (1x1 convolutions) to obtain rendering parameters.}
   \label{fig:cnn}
\end{figure*}

\begin{figure*}[ht] 
\centering
\includegraphics[width=0.7\textwidth]{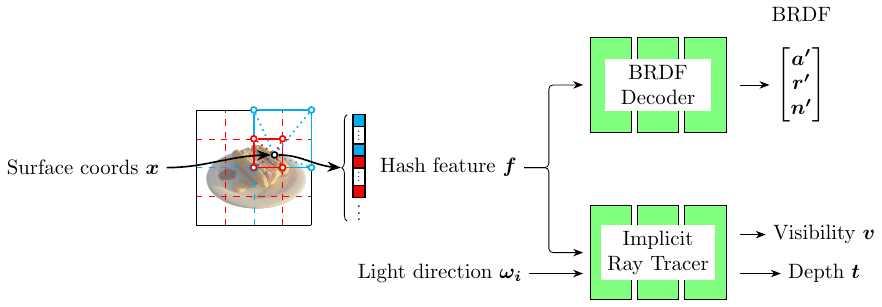}
\caption{Architecture of our proposed hash grid-based renderer for visibility modeling and indirect illumination.}\label{fig:indirect}
\end{figure*}

\section{Methodology}
\label{sec:method}

\subsection{Two-bounce Indirect Illumination Rendering}
We follow the physically-based rendering pipeline by evaluating Eqn. \ref{eqn:renderingeqn}. For this purpose, we model the primary surface point $\bm{x}$, BRDF parameters (albedo $a$, roughness $r$ from the simplified Disney model~\cite{simplifieddisney}, and surface normal $\bm{n}$), as well as visibility $V(\bm{x}, \bm{\omega}_i)$, with a deep neural network. To account for indirect illumination, secondary surface (bounce) point $\bm{x}'$ must be calculated to evaluate Eqn. \ref{eqn:renderingeqn} recursively. For light directions that correspond to shadow (occluded) rays, we adopt a strategy that models the expected depth $t$ from $\bm{x}$ to $\bm{x}'$ so that we can perform secondary ray tracing as $\bm{x}'=\bm{x}+t\cdot\bm{\omega}_i$. At $\bm{x}'$, we sample the light source again, collect only visible rays, and calculate the outgoing radiance $L'_o$. This value becomes the incoming illumination for the shadow rays. That is, the total incoming radiance is formulated as follows:

\begin{equation}
L(\bm{x}, \bm{\omega}_i)=V(x, \bm{\omega}_i) \cdot L_i(\bm{x}, \bm{\omega}_i) + (1-V(x, \bm{\omega}_i)) \cdot L'_o.
\end{equation}

\subsection{Proposed Model}
\paragraph{Overview}
The overall architecture and operation of our model are summarized in Fig. \ref{fig:overview}. The direct illumination renderer $G$ takes rays as input and outputs the material, normal, and primary surface coordinates ($\bm{x}$). The hash grid encoder $H$ produces a unified deep feature $f=H(\bm{x})$. The implicit ray tracer $I$ takes $f$ and the light direction, $\bm{\omega}_i$, as inputs to predict the hard visibility $v$ and the expected distance $t$ to the secondary surface point $\bm{x}'$ for each direction.

For indirect illumination, for each $\bm{\omega}_i$ where $v=0$ (shadow rays), we perform tracing to calculate the secondary surface point coordinates and input $\bm{x}'$ back into the hash encoder to obtain the secondary feature $f'=H(\bm{x}')$. The BRDF parameters at that location are calculated as $a',r',n'=B(f')$, and we then sample a new light direction $\bm{\omega}'_i$, to compute the secondary visibility $v'$. Only the $\bm{\omega}'_i$s where $v'=1$ (visible rays) are selected, and the rendering equation is evaluated to produce the indirect illumination, which is used as the incoming radiance for the shadow rays in the original equation at the primary surface coordinate $\bm{x}$.

\begin{figure*}[ht] 
\centering
\includegraphics[width=0.9\textwidth]{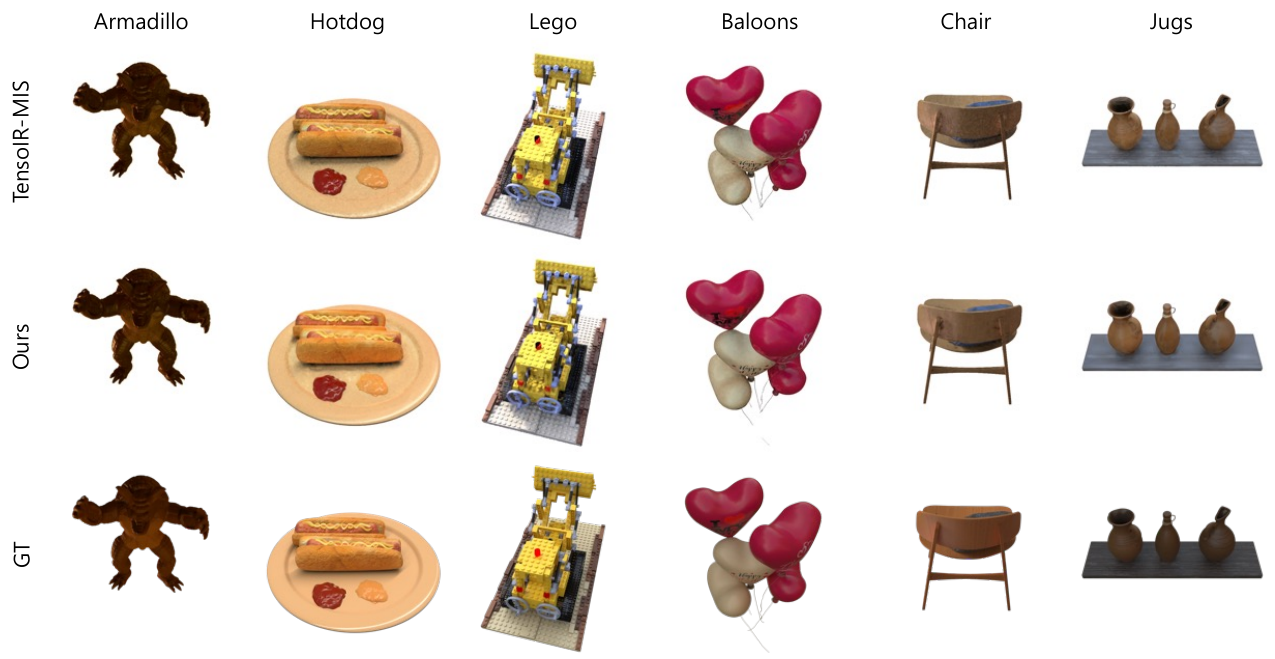}
\caption{Quantitative results on all 6 synthetic scenes. Environment lights and viewpoints are randomly selected for each scene.}\label{fig:main}
\end{figure*}

\paragraph{Direct Illumination Renderer}

Given a camera position $\bm{o}$, to perform direct illumination rendering, we propose a CNN-based renderer that combines light field-styled encoding (to reduce per-ray samples) with super-resolution (to reduce the number of rays required).

We compute the ray direction ($\bm{d}_{down}$) from $\bm{o}$ towards a downsampled pixel array, then broadcast $\bm{o}$ to concatenate with $\bm{d}_{down}$, to get the ray expression as $\bm{\gamma}=[\bm{o},\bm{d}_{down}]$ as the input. The task of the renderer is to generate full-resolution maps of material, normal, and primary surface coordinates. As illustrated in Fig. \ref{fig:cnn}, our architecture follows MobileR2L with a stacked ResMLP module for the encoder and three super-resolution modules for upsampling. From our initial experiments, we observe that naively using the MobileR2L's upsampling structure, where convolution layers are coupled with transposed convolution layers, leads to the development of checkerboard artifacts in the predicted maps~\cite{deconvartifact}, which is difficult to mitigate even with the addition of a smoothness loss. Consequently, we replace transposed convolution layers with a combination of bilinear upsampling and convolution layers~\cite{deconvartifact} and integrate the output skip structure from StyleGAN2~\cite{stylegan2} to achieve smoother outputs.

Similar to NeLF, we can obtain direct illumination with just one call to CNN forward, as opposed to methods such as NeRF, tensorf, and TensorIR, which require extensive computations like volume rendering. Specifically, previous methods had overhead associated with handling the varying number of valid sample points per ray, often requiring the use of bounding boxes or alpha masks for filtering before rewriting to the original tensor shape. In contrast, our method operates with a consistent input-output shape, thereby avoiding such issues.

\paragraph{Indirect Illumination Renderer}
To perform indirect illumination rendering quickly from the modeled surface point coordinates, as outlined in Fig. \ref{fig:indirect}, we propose a structure that learns unified feature through a hash grid encoder and calculates all necessary parameters (BRDF, visibility, and secondary ray depth) using only two very small MLPs. Firstly, the hash grid encoder $H$ is structured with multi-level tables as InstantNGP~\cite{instantngp}, accepting the surface coordinate $\bm{x}$ as input to output the feature $f$. Compared to the autoencoder or appearance voxel grid methods used in previous relightable NeRF models~\cite{invrender,tensoir}, this approach completes feature calculations solely through lookups and interpolation, thereby reducing per-ray sample points to the number of hash levels and significantly lightening per-sample computation. Moreover, the resulting feature is general to the scene, facilitating the attachment of multiple MLP decoders for easy multi-task learning.
Since BRDF parameters are quantities dependent only on location in the scene, we introduce a BRDF decoder $B$, an MLP that takes only $f$ as input to obtain these parameters. To reduce network call overhead, we treat $n$, $a$, $r$ as one concatenated output vector and later separate them for use in rendering.
Visibility and secondary depth depend on the shape of object and light direction, 
necessitating the introduction of an MLP implicit ray tracer $I$, which takes $f$ and direction $\bm{\omega}_i$ as inputs. This module performs two tasks:
1.	It predicts hard visibility via a hit test in the input direction.
2.	It predicts the expected depth in the input direction. Here, the output range is min-max normalized between predefined near and far values. The depth for visible rays should ideally be infinity, but as outputting such a value is practically challenging, we choose to clamp all outputs to 1 (corresponding to far) and then filter with visibility.
By leveraging such a compact architecture, our model achieves efficient parameter computation while maintaining rendering effectiveness across various tasks.

\subsection{Training Procedure}
We utilize a teacher model, TensoIR which is pre-trained according to its original procedure. To train our direct illumination renderer, we sample 10,000 random poses 
\footnote{The large number of random poses is analogous to observations in light field network, as around 100 original training views are insufficient for accurate prediction from arbitrary views.}
from the hemisphere where the training set's camera poses are located. During this process, we include a rejection step to ensure that there is no overlap with the training or test views.

For every random and training pose, we generate rays and run TensoIR to extract albedo, roughness, surface normal, surface coordinate, and object mask at full resolution for each pose. Our renderer uses these maps as pseudo or training data.

From the training set, we select random rays to input into TensoIR and retrieve the BRDF parameters for the corresponding pixel, then employ stratified sampling to generate random incoming light directions $\bm{\omega}_i$. Volume rendering is performed on the density voxel grid only for surface normals that are front-facing, to calculate the transmittance $T$ and the expected $z$ value (depth) $t$. $T$ is binarized with a threshold of 0.5. This process results in a batch containing three types of pseudo data.

In order to improve training speed, we primarily try to exploit pre-computation because the teacher model is computationally expensive, and extracting online would make the training time for our model excessively long. 
One exception is for training the hash renderer where we adopt an online approach. For visibility distillation, it is essential to train in numerous random directions, which would be impractical to pre-compute. 

The training loop for the hash renderer consists of an optimization process. First, we obtain the hash feature as $f=H(\bm{x})$.
We compute $\bm{n}, a, r=B(f)$, after which the normal is normalized ($n'=n/||n||$), the albedo is processed as an RGB image ($a'=\sigma(a)$), and the roughness values are constrained between 0.09 and 1.0 ($r=0.09 + \sigma(r) * 0.91$). 
We minimize the L1 loss between these values and the teacher output.
We compute visibility $v$ and depth $t=I(f, \bm{\omega}_i)$, applying sigmoid to the visibility prediction to map it to the range [0, 1], and using min-max normalization on the depth with near and far values to also map it to [0, 1]. These are then trained using binary cross entropy loss and L1 loss, respectively.

\section{Experiments}
\label{sec:experiments}

\subsection{Experimental Settings}
We employ two synthetic datasets, TensoIR-synthetic and InvRender-synthetic, and experiment on 3 scenes from each. The image resolution is fixed to 800$\times$800. For our CNN-based direct illumination renderer, we mostly follow the settings from MobileR2L~\cite{mr2l}. Our hash encoder has 14 levels and the table size is set to $2^{17}$ with a feature dimension of 2 for each entry. The minimum and maximum resolutions are set to 16 and 131,072, respectively. The BRDF decoder and implicit ray tracer both have 3 layers in total with hidden dimensions set to 64.

For efficient all-frequency rendering, we apply MIS~\cite{veach} where a light importance ray and a GGX importance ray are sampled for each primary surface point. The probability distribution function for the rays is computed with Veach's balance heuristic~\cite{veach}. The GGX rays are drawn from visible normal distribution function~\cite{vndf} to further increase sampling efficiency. Following \cite{nvdiffrecmc}, we use roughness remapping as $\alpha={Roughness}^2, k=\alpha^2$. SVGF~\cite{svgf}, one of the most popular non-parametric denoising filters, operates on the raw MC rendering results to greatly reduce pixel color variance. Since our model is not targeting video generation, all temporal elements of SVGF are removed.

All the other details not mentioned here can be found in the supplementary material.

\subsection{Teacher Model}
We utilize TensoIR~\cite{tensoir} as a teacher model. Using the authors' code to reproduce results, we identified an issue with roughness overestimation. Therefore, instead of the commonly used $\alpha=r^2$ mapping in Specular BRDF calculations, we use a version that improves this by replacing the Fresnel function with the Schlick approximation and the microfacet visibility function with the Correlated Smith approximation. While TensoIR performs rendering with MC using stratified sampling, it learns the light source through approximation using SG; hence, the optimal reparameterization settings appear to differ from the original. Without taking these measures, there's an issue where the microfacet normal is sampled too widely, almost failing to represent specular highlights.

Furthermore, the original implementation of TensoIR only supports light importance rays with inefficient sampling routine. This leads to poor representation of specular colors, and the quantity of rays that can be processed in parallel is limited, causing the total latency to be up to 20 seconds. To address this, we re-implement the rendering pipeline of TensoIR to make a highly optimized version, named TensoIR-MIS, that includes MIS with SVGF and applies optimization features provided by PyTorch as much as possible. Note that all results from TensoIR-MIS are obtained by manually tuning the chunk size to maximize GPU core and memory utilization, and we report the fastest achievable results.

\subsection{Student Model}
For our CNN renderer, each module inside the residual MLP has an identical architecture to the initial conv-norm-act layer. The output maps of an SR block are bilinearly upsampled and added to the maps from the next SR block until the final resolution is reached.
The first convolution layer and all convolution layers in the residual MLP module each have a channel of 256 and a kernel size of 1. The channel size gets reduced by 2 times for each SR block (128 for the first block, 64 for the second, and so on). The residual MLP has 28 identical conv-norm-act layers inside.

Further details and all configuration files will be made public in our GitHub repository.

\begin{table*}[h]
\centering
\caption{Quantitative comparison on 4 scenes from TensoIR-synthetic dataset. Our model exhibits similar PSNR, SSIM, and LPIPS metrics, while having up to 53$\times$ faster rendering speed as shown in Table \ref{tab:latency}. All quality metrics are measured with respect to ground truth image.}
\label{tab:tensoirsynth}
\resizebox{\textwidth}{!}{%
\begin{tabular}{@{}lcccccccccccccccc@{}}
\toprule
\multicolumn{1}{c}{\multirow{2}{*}{Model}} & \multicolumn{3}{c}{armadillo}              & \multicolumn{3}{c}{ficus}                  & \multicolumn{3}{c}{hotdog}                 & \multicolumn{3}{c}{lego}                   & \multicolumn{3}{c}{Average}                & Speed \\ \cmidrule(l){2-17} 
\multicolumn{1}{c}{}                       & PSNR  & SSIM  & LPIPS                      & PSNR  & SSIM  & LPIPS                      & PSNR  & SSIM  & LPIPS                      & PSNR  & SSIM  & LPIPS                      & PSNR  & SSIM  & LPIPS                      & FPS   \\ \midrule
TensoIR-MIS (16, d)                        & 32.77 & 0.965 & \multicolumn{1}{c|}{0.067} & 24.83 & 0.947 & \multicolumn{1}{c|}{0.073} & 28.13 & 0.935 & \multicolumn{1}{c|}{0.110} & 26.06 & 0.897 & \multicolumn{1}{c|}{0.111} & 27.95 & 0.936 & \multicolumn{1}{c|}{0.090} & 0.82  \\
TensoIR-MIS (32, d)                        & 32.91 & 0.968 & \multicolumn{1}{c|}{0.065} & 24.89 & 0.948 & \multicolumn{1}{c|}{0.072} & 28.84 & 0.944 & \multicolumn{1}{c|}{0.099} & 26.63 & 0.912 & \multicolumn{1}{c|}{0.099} & 28.32 & 0.943 & \multicolumn{1}{c|}{0.084} & 0.72  \\
TensoIR-MIS (64, d)                        & 32.97 & 0.970 & \multicolumn{1}{c|}{0.064} & 24.92 & 0.949 & \multicolumn{1}{c|}{0.071} & 29.19 & 0.949 & \multicolumn{1}{c|}{0.091} & 26.89 & 0.921 & \multicolumn{1}{c|}{0.089} & 28.49 & 0.947 & \multicolumn{1}{c|}{0.079} & 0.56  \\
Ours (16, d)                               & 32.60 & 0.963 & \multicolumn{1}{c|}{0.073} & 24.59 & 0.944 & \multicolumn{1}{c|}{0.079} & 27.67 & 0.927 & \multicolumn{1}{c|}{0.120} & 24.62 & 0.868 & \multicolumn{1}{c|}{0.136} & 27.37 & 0.926 & \multicolumn{1}{c|}{0.102} & 69.44 \\
Ours (32, d)                               & 32.74 & 0.966 & \multicolumn{1}{c|}{0.071} & 24.65 & 0.945 & \multicolumn{1}{c|}{0.078} & 28.35 & 0.935 & \multicolumn{1}{c|}{0.111} & 25.03 & 0.879 & \multicolumn{1}{c|}{0.128} & 27.69 & 0.931 & \multicolumn{1}{c|}{0.097} & 47.62 \\
Ours (64, d)                               & 32.79 & 0.967 & \multicolumn{1}{c|}{0.071} & 24.68 & 0.946 & \multicolumn{1}{c|}{0.078} & 28.69 & 0.939 & \multicolumn{1}{c|}{0.105} & 25.22 & 0.886 & \multicolumn{1}{c|}{0.121} & 27.85 & 0.935 & \multicolumn{1}{c|}{0.094} & 28.25 \\ \midrule
TensoIR-MIS (16, d+i)                      & 32.83 & 0.966 & \multicolumn{1}{c|}{0.068} & 24.83 & 0.948 & \multicolumn{1}{c|}{0.073} & 29.96 & 0.934 & \multicolumn{1}{c|}{0.110} & 25.66 & 0.893 & \multicolumn{1}{c|}{0.114} & 28.32 & 0.935 & \multicolumn{1}{c|}{0.091} & 0.42  \\
TensoIR-MIS (32, d+i)                      & 32.96 & 0.969 & \multicolumn{1}{c|}{0.065} & 24.89 & 0.948 & \multicolumn{1}{c|}{0.072} & 30.41 & 0.942 & \multicolumn{1}{c|}{0.101} & 26.03 & 0.907 & \multicolumn{1}{c|}{0.103} & 28.57 & 0.942 & \multicolumn{1}{c|}{0.085} & 0.27  \\
TensoIR-MIS (64, d+i)                      & 33.02 & 0.970 & \multicolumn{1}{c|}{0.064} & 24.92 & 0.949 & \multicolumn{1}{c|}{0.071} & 30.50 & 0.946 & \multicolumn{1}{c|}{0.094} & 26.22 & 0.915 & \multicolumn{1}{c|}{0.094} & 28.67 & 0.945 & \multicolumn{1}{c|}{0.081} & 0.16  \\
Ours (16, d+i)                             & 32.69 & 0.964 & \multicolumn{1}{c|}{0.073} & 24.63 & 0.945 & \multicolumn{1}{c|}{0.079} & 29.29 & 0.930 & \multicolumn{1}{c|}{0.118} & 24.98 & 0.870 & \multicolumn{1}{c|}{0.134} & 27.90 & 0.927 & \multicolumn{1}{c|}{0.101} & 36.23 \\
Ours (32, d+i)                             & 32.83 & 0.966 & \multicolumn{1}{c|}{0.071} & 24.68 & 0.945 & \multicolumn{1}{c|}{0.078} & 29.90 & 0.937 & \multicolumn{1}{c|}{0.110} & 25.31 & 0.881 & \multicolumn{1}{c|}{0.127} & 28.18 & 0.932 & \multicolumn{1}{c|}{0.097} & 22.03 \\
Ours (64, d+i)                             & 32.88 & 0.968 & \multicolumn{1}{c|}{0.071} & 24.71 & 9.946 & \multicolumn{1}{c|}{0.078} & 30.15 & 0.941 & \multicolumn{1}{c|}{0.104} & 25.48 & 0.888 & \multicolumn{1}{c|}{0.120} & 28.31 & 0.936 & \multicolumn{1}{c|}{0.093} & 11.92 \\ \bottomrule
\end{tabular}%
}
\end{table*}

\begin{table*}[h]
\centering
\caption{Quantitative comparison on 3 scenes from InvRender-synthetic dataset. The hotdog scene is excluded as it also exists in the TensoIR-synthetic dataset. In all scenes, ours even outperforms TensoIR with much faster rendering speed.}
\label{tab:invrendersynth}
\resizebox{0.8\textwidth}{!}{%
\begin{tabular}{lccc|ccc|ccc}
\hline
\multicolumn{1}{c}{\multirow{2}{*}{Model}} & \multicolumn{3}{c|}{baloons} & \multicolumn{3}{c|}{chair} & \multicolumn{3}{c}{jugs} \\ \cline{2-10} 
\multicolumn{1}{c}{}                       & PSNR     & SSIM    & LPIPS   & PSNR    & SSIM    & LPIPS  & PSNR   & SSIM   & LPIPS  \\ \hline
TensoIR-MIS (32, d)                         & 28.17    & 0.930   & 0.105   & 26.82   & 0.938   & 0.092  & 24.88  & 0.955  & 0.058  \\
TensoIR-MIS (64, d)                        & 28.75    & 0.943   & 0.095   & 27.13   & 0.952   & 0.083  & 24.89  & 0.960  & 0.051  \\
Ours (32, d)                               & 29.27    & 0.949   & 0.091   & 27.37   & 0.946   & 0.094  & 25.31  & 0.959  & 0.063  \\
Ours (64, d)                               & 29.28    & 0.952   & 0.088   & 27.35   & 0.949   & 0.092  & 25.24  & 0.960  & 0.062  \\ \hline
TensoIR-MIS (32, d+i)                       & 28.42    & 0.933   & 0.103   & 27.07   & 0.918   & 0.129  & 24.54  & 0.954  & 0.058  \\
TensoIR-MIS (64, d+i)                      & 28.68    & 0.945   & 0.094   & 27.01   & 0.922   & 0.121  & 24.49  & 0.958  & 0.051  \\
Ours (32, d+i)                             & 29.18    & 0.952   & 0.090   & 27.23   & 0.948   & 0.092  & 24.91  & 0.956  & 0.064  \\
Ours (64, d+i)                             & 29.08    & 0.955   & 0.087   & 27.12   & 0.950   & 0.090  & 24.83  & 0.958  & 0.062  \\ \hline
\end{tabular}%
}
\end{table*}

\subsection{Results on Synthetic Scenes}

Tables \ref{tab:tensoirsynth} and \ref{tab:invrendersynth} compare ours with the teacher models. As the tables show, ours offer comparable rendering quality to the teacher models on the synthetic scenes. The visualization of rendered results is presented in Fig. \ref{fig:main}.

\subsection{Runtime Analysis}

\begin{table*}[h]
\centering
\caption{Latency breakdown results on TensoIR-MIS and ours. Model Op., Vis (+id), Render, Dnsr mean the time required for i) obtaining rendering parameters for direct illumination rendering, ii) computing visibility and running all procedures for indirect illumination rendering, iii) evaluating rendering equation, and iv) operating denoising filter, respectively. To evaluate the speedup ratio, the FPS of our model is divided by TensoIR-MIS' FPS on the matched spp.}
\label{tab:latency}
\resizebox{0.8\textwidth}{!}{%
\begin{tabular}{lccccccc}
\hline
Model                  & Model Op. & Vis    & Render & Dnsr & Latency (ms) & FPS   & Speedup \\ \hline
TensoIR-MIS (16, d)    & 1008.3    & 187.2  & 19.7   & 3.3  & 1218.5       & 0.82  & N/A     \\
TensoIR-MIS (32, d)    & 1008.5    & 363.6  & 21.2   & 3.3  & 1396.6       & 0.72  & N/A     \\
TensoIR-MIS (64, d)    & 1009.2    & 741.5  & 39.2   & 3.3  & 1793.2       & 0.56  & N/A     \\
Ours (16, d)           & 4.3       & 4.6    & 2.2    & 3.3  & 14.4         & 69.44 & 84.62   \\
Ours (32, d)           & 4.3       & 9.2    & 4.3    & 3.2  & 21.0         & 47.62 & 66.50   \\
Ours (64, d)           & 5.3       & 18.4   & 8.4    & 3.3  & 35.4         & 28.25 & 50.66   \\ \hline
TensoIR-MIS (16, d+id) & 1011.2    & 1325.3 & 17.2   & 3.3  & 2357         & 0.42  & N/A     \\
TensoIR-MIS (32, d+id) & 1009.2    & 2605.5 & 19.2   & 3.3  & 3637.2       & 0.27  & N/A     \\
TensoIR-MIS (64, d+id) & 1011.6    & 5201.2 & 25.8   & 3.3  & 6241.9       & 0.16  & N/A     \\
Ours (16, d+id)        & 5.4       & 16.3   & 2.6    & 3.3  & 27.6         & 36.23 & 85.40   \\
Ours (32, d+id)        & 5.3       & 32.4   & 4.4    & 3.3  & 45.4         & 22.02 & 80.11   \\
Ours (64, d+id)        & 5.8       & 64.9   & 9.7    & 3.5  & 83.9         & 11.92 & 74.40   \\ \hline
\end{tabular}%
}
\end{table*}

As listed in Table \ref{tab:latency}, our model is up to 85.40 times faster than the TensoIR-MIS model with the same spp setting. We measure all latency values with a Nvidia A100 40GB SXM2 GPU with exactly the same hardware and software environment. Since TensoIR-MIS requires a large amount of VRAM per pixel even with our re-implementation, it must process chunks of 4k to 16k rays at a time and iterate until all pixel colors are evaluated. On the other hand, ours can even process 2 (with spp of 64) or 4 (with spp of 32) images at once, due to the efficiently baked model architecture. This explains the huge speedup ratio of about 500 for model operation. For visibility, TensoIR-MIS has to perform volume rendering on its density voxel grid for each secondary ray, while ours does lookups on the hash table and runs tiny MLPs, which are easy to heavily exploit the parallelism of modern GPUs.

\subsection{Results on Real Scenes}
We also conducted experiments on real scenes. OpenIllumination benchmark~\cite{openillumination} is employed and we select 4 random scenes (bird, metal bucket, pumpkin, and sponge). The $013$ lighting pattern is used as training light condition and $003$, $006$, $009$, and $012$ lighting patterns are selected for relighting evaluation. The patterns are originally provided as sets of SGs and we render them to equirectangular HDR environment maps. To train a TensoIR-MIS model, we use the configurations found in the original OpenIllumination codebase except that all images are downsampled by a factor of 3.32 and center-cropped to have a final resolution of 800 by 800. 

\begin{figure}[h] 
\centering
\includegraphics[width=0.7\columnwidth]{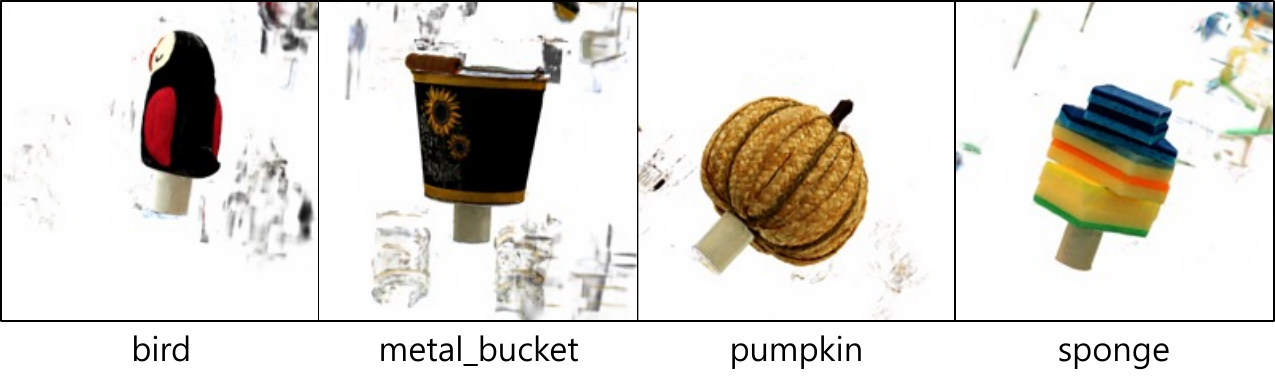}
\caption{Visualization of albedo maps with floater artifacts extracted from TensoIR-MIS teachers.}\label{fig:floater}
\end{figure}

A lot of floater artifacts are observed for all scenes from the inverse rendering results of TensoIR-MIS as shown in Fig. \ref{fig:floater}. They also appear on the pseudo data since there is no ground truth object mask for the corresponding camera pose. Thus, they significantly hinder the learned representation of our direct illumination renderer which has to perform volume rendering in a fully implicit way. We observe that our CNN renderer faithfully tries to reconstruct all such floaters, which appears to deteriorate its multi-view synthesis capability.

\begin{figure}[h] 
\centering
\includegraphics[width=0.7\columnwidth]{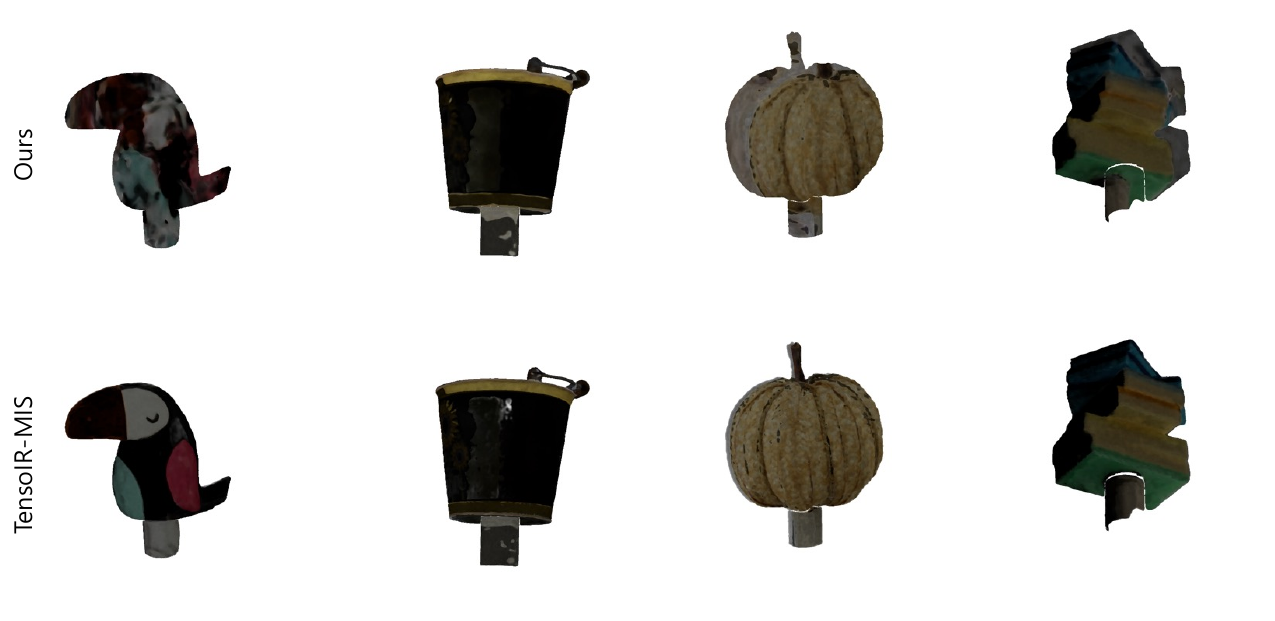}
\caption{Quantitative comparison on 4 scenes from OpenIllumination benchmark. The spp is set to 32 and both direct and indirect illumination are rendered. The test view and environment map are randomly selected.}\label{fig:openillum}
\end{figure}

Even with object masks applied to the final rendered RGB map, the CNN renderer behaves unexpectedly so that as visualized in Fig \ref{fig:openillum}, details are abruptly erased (especially for bird) and the rendered images are as if they were from shifted camera pose (especially for pumpkin). It is clear that a method to remove floaters from the teacher model (e.g., stronger sparsity constraint on TensoIR's voxel grids) is imperative to making our model perform as designed, however, we leave it as future work.

\begin{figure}[h] 
\centering
\includegraphics[width=0.7\textwidth]{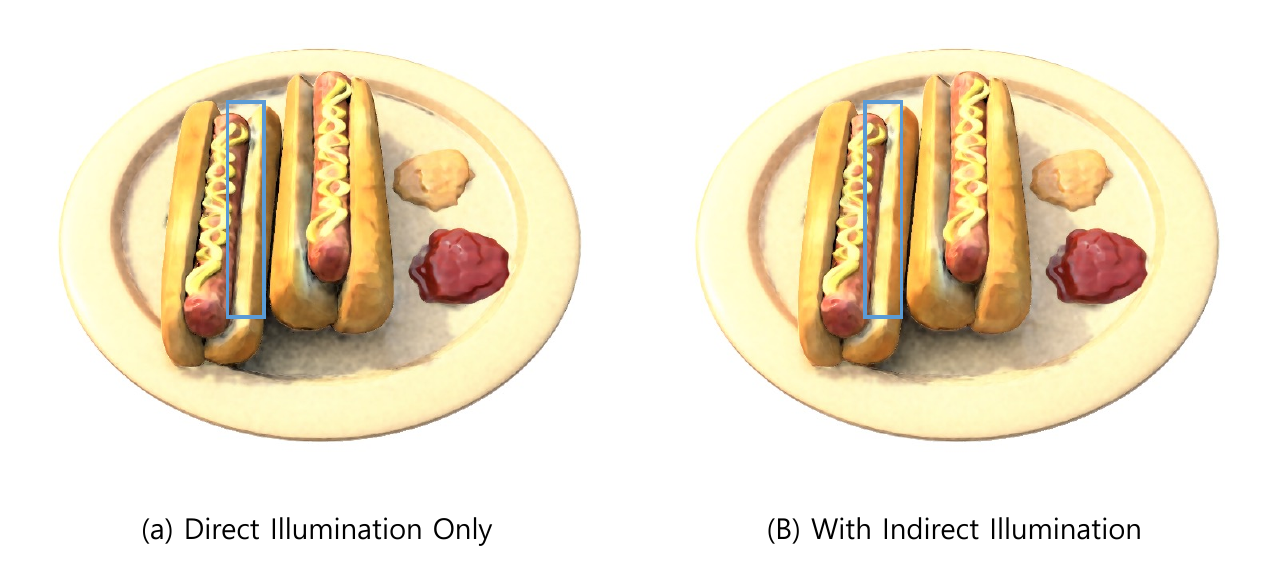}
\caption{Qualitative result showing the effect of rendering indirect illumination. As highlighted by the blue boxes, the narrow regions with steep geometry change can benefit from 2-bounce rays (best viewed when zoomed in).}\label{fig:indirecteffect}
\end{figure}

\subsection{Ablation Studies}

\begin{figure}[h] 
\centering
\includegraphics[width=0.7\textwidth]{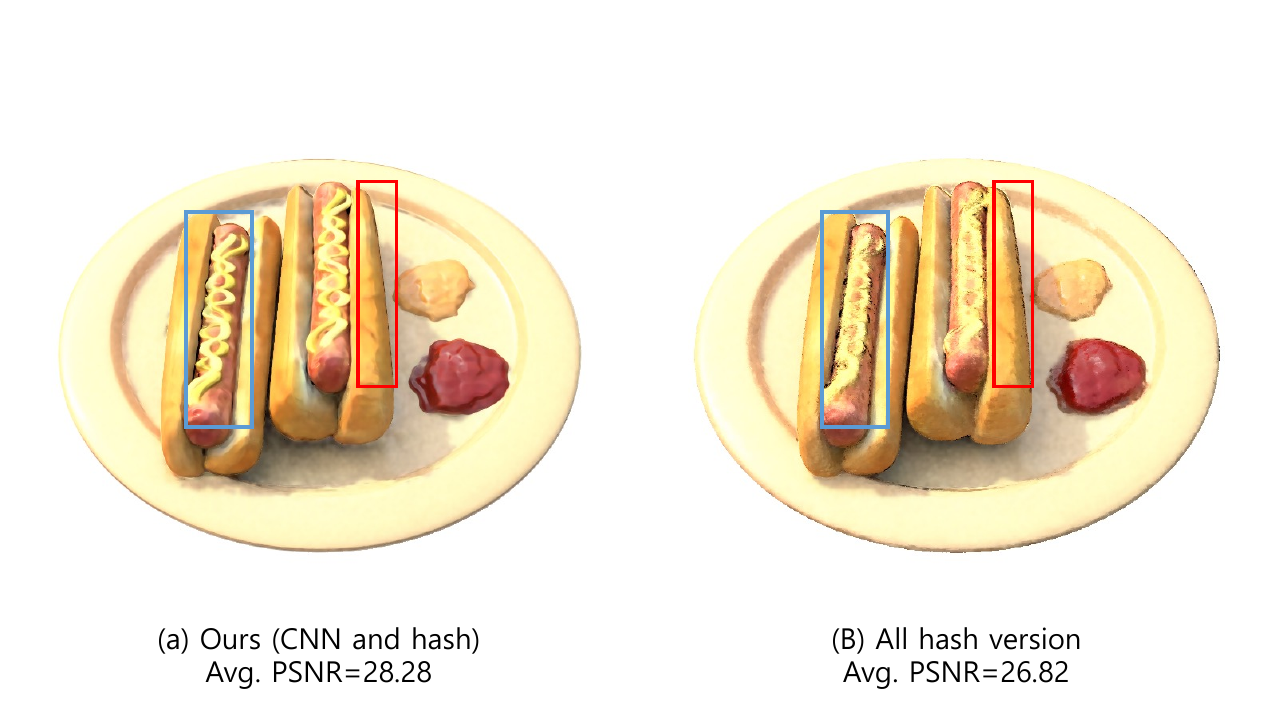}
\caption{Qualitative result on an ablation study regarding the architectural variation of our model. As highlighted by the colored boxes, details, especially near edges and complex regions (mustard sauce on sausages in this case), are severely degraded in the case of all hash version.}\label{fig:indirecteffect}
\end{figure}

\paragraph{Effect of Indirect Illumination Rendering}
Our proposed hash renderer can independently perform rendering when given primary surface coordinates. If we constrain the role of our CNN renderer to depth prediction, a variant of our model can be created that performs all operations, excluding primary ray casting, via hashing. However, experimental results reveal a significant disturbance in the predicted BRDF values due to slight depth errors, leading to a degradation in the quality of direct illumination. Particularly evident in the portion of Fig. \ref{fig:indirecteffect} highlighted by blue boxes, the hit test results on rapid geometry changes and narrow areas are mostly identified as occlusions, leading to unwanted dark pixels and a substantial drop in PSNR.

\begin{table}[ht]
\centering
\caption{Ablation on the size of our direct illumination renderer. The results are from TensoIR-synthetic benchmark. In Ours-S, all channel values (widths) for the direct illumination renderer are halved. }
\label{tab:smallcnn}
\resizebox{0.7\textwidth}{!}{%
\begin{tabular}{@{}lcc|lcc@{}}
\toprule
\multicolumn{1}{c}{Model} & Avg. PSNR & FPS   & \multicolumn{1}{c}{Model} & Avg. PSNR & FPS   \\ \midrule
Ours-S (16, d)            & 26.75     & 79.18 & Ours-S (16, d+i)          & 27.30     & 39.49 \\
Ours (16, d)              & 27.37     & 69.44 & Ours (16, d+i)            & 27.90     & 36.23 \\
Ours-S (32, d)            & 27.05     & 52.00 & Ours-S (32, d+i)          & 27.56     & 23.14 \\
Ours (32, d)              & 27.69     & 47.62 & Ours (32, d+i)            & 28.18     & 22.03 \\
Ours-S (64, d)            & 27.20     & 30.10 & Ours-S (64, d+i)          & 27.67     & 12.27 \\
Ours (64, d)              & 27.85     & 28.25 & Ours (64, d+i)            & 28.31     & 11.92 \\ \bottomrule
\end{tabular}%
}
\end{table}

\paragraph{Model Size}
In Table \ref{tab:smallcnn}, we examine how the size of our direct illumination renderer affects the quality and speed of our whole model. We observe a tradeoff between a loss in averaged PSNR up to 0.7 and a gain in FPS up to 10. However, the speedup is less visible when the spp value is increased or indirect illumination is rendered. This is because the latency of the CNN renderer does not change a lot given the same number of primary rays.


\begin{figure}[h] 
\centering
\includegraphics[width=0.5\columnwidth]{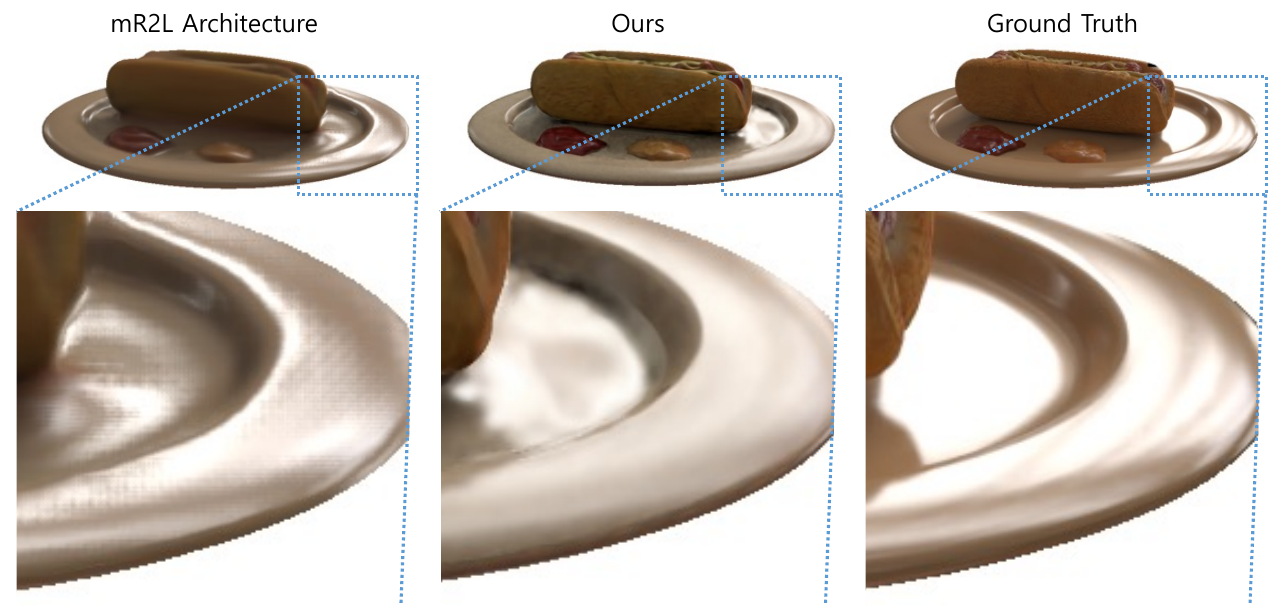}
\caption{Visualization of checkerboard artifacts generated by the original mR2L architecture.}\label{fig:mr2lcompare}
\end{figure}
\paragraph{Architecture of the CNN Renderer}
As mentioned in the main paper, we qualitatively show the checkerboard artifacts generated by the transposed convolution-based upsampling layer used in the original MobileR2L architecture in Fig. \ref{fig:mr2lcompare}. Only direct illumination is rendered. Though they don't critically impact the quality metrics and are barely visible when zoomed out, one can easily perceive them in the original resolution.

\section{Limitations and Future Works}
The main limitation of our proposed model is that it cannot correct problems coming from the teacher model. For instance, we use a fixed Fresnel value of 0.04 as in TensoIR, which limits the model's capability of modeling metallic surfaces. Ours follows a lot of existing relightable NeRF models including TensoIR to apply hard surface assumption, which makes it impossible to correctly model refraction and transparency. As clearly shown in Fig. \ref{fig:openillum}, when supervision from the teacher has low quality, ours can generate unsatisfactory results.

The CNN and hash-based renderers can be further optimized by extensive hyperparameter search, quantization, pruning, etc. We leave the optimizations as future work.

It will be an interesting topic to extend our baked renderers to enable dynamic relightable neural rendering by a carefully designed strategy to make them learn temporal information.

\section{Conclusion}
\label{sec:conclusion}
{
We proposed a real-time relighting method that supports both direct and indirect illumination under unseen lighting conditions.
Given a pre-trained teacher model, our proposed CNN-based renderer, which builds upon an existing method, MobileR2L judiciously exploits the GAN architecture to resolve the problem of the existing method while enabling high-quality direct illumination.
Our proposed lightweight hash grid-based renderer consists of multi-scale hash grid and two small MLPs (implicit ray tracer and BRDF decoder) and serves direct illumination by providing visibility and indirect illumination by predicting the coordinate and BRDF parameters of the secondary surface point.
Our extensive experiments evaluate the abilities of direct and indirect illumination and show that our proposed method offers up to orders of magnitude faster rendering speed with comparable rendering quality to the teacher models. 
}


%
%
\bibliographystyle{splncs04}
\bibliography{main}
\end{document}


\title{Supplementary Material for Baking Relightable NeRF for Real-time Direct/Indirect Illumination Rendering}

\maketitle

\section{More Ablation Studies}

\subsection{SG-based Direct Illumination}
\begin{figure}[h] 
\centering
\includegraphics[width=0.7\columnwidth]{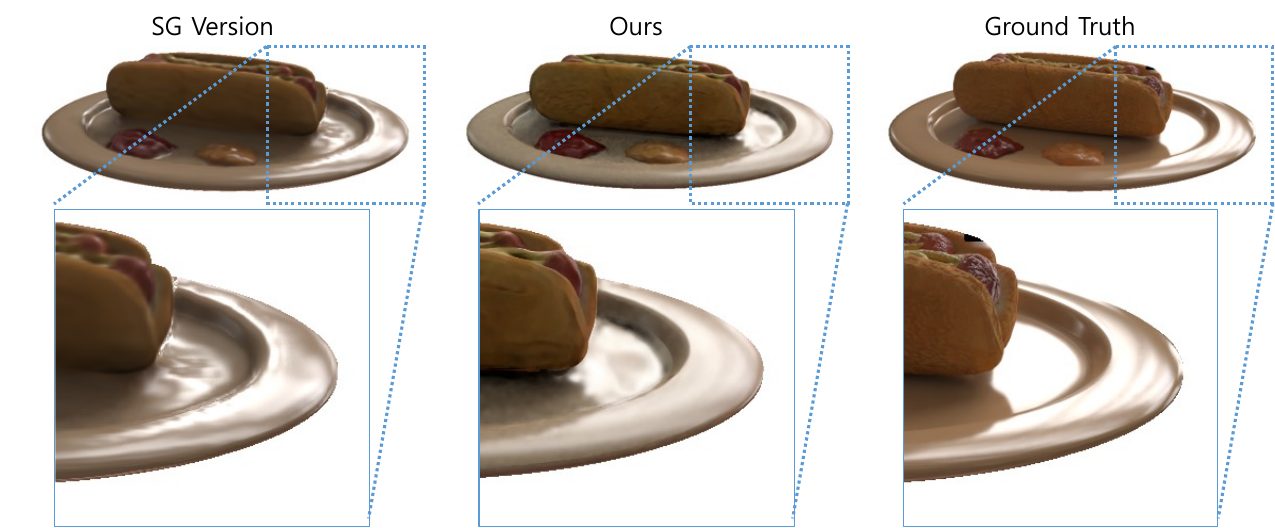}
\caption{Qualitative comparison on our model and its SG variant.}\label{fig:sgcompare}
\end{figure}
InvRender~\cite{invrender}, which is based on SG lighting, can also serve as a teacher for our model with the same experimental setting as TensoIR.

Since an SG light source makes a non-zero contribution to all incoming light directions that are forward-facing with surface normal, it is impossible to employ hard shadow and instead required to \textit{estimate} soft visibility at test time. InvRender~\cite{invrender} accomplishes this by sampling random directions near the lobe of an SG with their sampling range adjusted by the sharpness of the SG and evaluating the MC integral. 

We modify our direct illumination renderer to make an SG variant that can learn from an InvRender teacher. The architecture of our direct illumination renderer is directly adopted with some minor modifications for visibility modeling. We directly distill InvRender's specular visibility ratio and follow the MC estimation pipeline for diffuse visibility.

Though this SG variant performs better than ours in terms of PSNR with a gap of approximately 1 on the hotdog scene, it comes at a price of a lot of distortions in the shape of shadows and specular lights, as shown in Fig. \ref{fig:sgcompare}. Furthermore, with soft visibility enforced, we find it tricky to model indirect illumination on this variant.

\subsection{Effect of Roughness Remapping on Teacher}
\begin{figure}[h] 
\centering
\includegraphics[width=0.7\columnwidth]{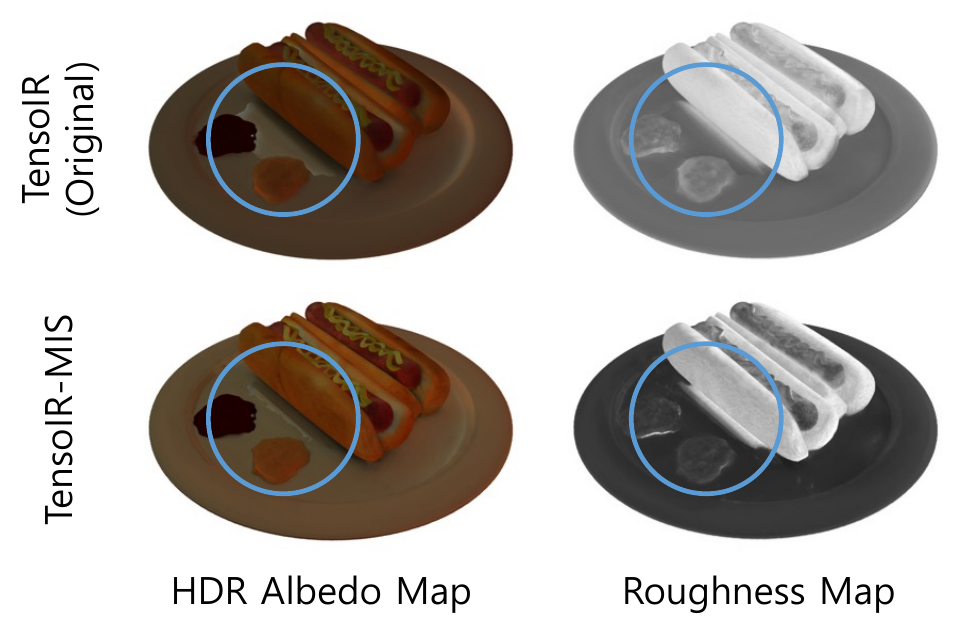}
\caption{Comparing our modified TensoIR-MIS to the original TensoIR's rendering setting. As highlighted with blue circles, the BRDF decomposition on a part of the dish is enhanced. For roughness, a darker color means a lower value, leading to stronger specular lights.}\label{fig:remapcompare}
\end{figure}
InvRender~\cite{invrender}, which is based on SG lighting, can also serve as a teacher for our model with the same experimental setting as TensoIR.
The overall quality of our model is upper-bounded by the teacher, which makes it an important step for us to improve TensoIR to the best. As Fig. \ref{fig:remapcompare} shows, if a widely used roughness remapping $\alpha=Roughness^2$ and uncorrelated Smith approximation is adopted for masking-shadowing function, two problems appear on the teacher's quality. Firstly, large mirror-like artifacts in decomposed BRDF parameters are observed. They don't affect the reconstruction results for the training dataset with a single lighting condition, however, they become clearly visible under novel lighting conditions. We can also interpret this behavior as overfitting to the training lights and our modification is partially helpful for generalization. Secondly, TensoIR tends to overestimate the roughness value which makes the final rendering poor in terms of specular lighting, even with the advanced rendering methods we add to its original implementation.


\section{Detailed Experimental Settings}
\subsection{Model Specifications}
All TensoIR-MIS models have identical specifications to the original TensoIR.



\subsection{Training Details}
\paragraph{CNN Renderer}
The camera poses to extract pseudo-data are randomly drawn from the upper hemisphere. 64 random poses, which are equivalent to predicting 64 pairs of all rendering parameters (albedo, normal, roughness, and surface coordinate), are drawn for each minibatch to train our direct illumination renderer. Pseudo-data extraction and training take approximately 1 and 8 hours, respectively.
Adam optimizer is adopted with the initial learning rate of $5e-4$ and the learning rate is cosine decayed throughout the training steps (100k steps in total).

\paragraph{Hash Renderer}
For each primary surface point, 1,024 secondary rays are cast by stratified sampling on the whole sphere, and those that are not forward-facing with surface normal are excluded from the loss calculation. The near and far values for secondary ray tracing (that is conducted on the density voxel grid of TensoIR) are set to 0.05 and 1.5, respectively, with 96 samples on each ray. This means that all secondary depth values are clipped to a range of [0.05, 1.5]. 2,048 random rays from the training dataset are drawn for each minibatch and the training includes 80k updates that take approximately 1.5 hours. Adam optimizer is adopted with the initial learning rate of $2e^{-3}$ and we follow the learning rate decay scheme of TensoIR. A weight decay of $1e^{-6}$ is applied to all MLPs in our indirect illumination renderer.

\subsection{Rendering Details}
We follow nerfactor~\cite{nerfactor} and TensoIR~\cite{tensoir} to calibrate the predicted albedo maps by the global albedo rescaling factor. For TensoIR-synthetic and InvRender-synthetic benchmark, the factor is calculated for RGB channels and red channel, respectively. For OpenIllumination~\cite{openillumination}, we follow the authors not to apply albedo calibration.

For indirect illumination rendering, we cast 8 random rays from each secondary surface point via stratified sampling.

The clip-space depth gradient map required to operate SVGF \cite{svgf,nvdiffrecmc} is approximated by offsetting the predicted depth map (derived from the direct illumination renderer's primary surface coordinate map) vertically and horizontally and calculating the numerical difference to the original value.

\subsection{Computing Environment}
For pre-training TensoIR-MIS models, a single RTX3090 GPU is required. Extracting pseudo datasets from them is parallelized to 8 GPUs.
We make use of 8 RTX3090 or RTX4090 GPUs to train our direct illumination renderer, and a single RTX3090 for our indirect illumination renderer.

For evaluation, we measure all metrics and latency values with an Nvidia A100 40GB SXM2 GPU with exactly the same hardware and software environment. The versions of GPU driver and CUDA libraries are 525.105.17-server and 12.1, respectively. For python environment, an NGC PyTorch container \footnote{We use 23.06 version from \url{https://catalog.ngc.nvidia.com/orgs/nvidia/containers/pytorch}.} is deployed on Ubuntu 20.04.

\pagebreak

%
%
\bibliographystyle{splncs04}
\bibliography{main}